\documentclass[submission,copyright,creativecommons]{eptcs}
\providecommand{\event}{FEAR 2025} 

\usepackage{iftex}

\ifpdf
  \usepackage{underscore}         
  \usepackage[T1]{fontenc}        
\else
  \usepackage{breakurl}           
\fi

\usepackage{amsmath,amsfonts}
\usepackage[obeyFinal]{todonotes}
\usepackage{xspace}

\newcommand{\asf}{\ensuremath{\textsf{a}}\xspace}

\newcommand{\Fbf}{\ensuremath{\mathbf{F}}\xspace}
\newcommand{\Obf}{\ensuremath{\mathbf{O}}\xspace}
\newcommand{\Pbf}{\ensuremath{\mathbf{P}}\xspace}

\newcommand{\Fmc}{\ensuremath{\mathcal{F}}\xspace}
\newcommand{\Omc}{\ensuremath{\mathcal{O}}\xspace}
\newcommand{\Smc}{\ensuremath{\mathcal{S}}\xspace}

\newcommand{\From}{\ensuremath{\textsf{From}}\xspace}

\title{NAEL: Non-Anthropocentric Ethical Logic}
\author{Bianca Maria Lerma \qquad \qquad Rafael Pe\~naloza
\institute{University of Milano-Bicocca, Milan, Italy}
\email{biancalerma99@gmail.com, rafael.penaloza@unimib.it}
}
\def\titlerunning{NAEL}
\def\authorrunning{B.M. Lerma \& R. Pe\~naloza}
\begin{document}
\maketitle

\begin{abstract}
We introduce NAEL (Non-Anthropocentric Ethical Logic), a novel ethical framework for artificial agents grounded in active inference and symbolic reasoning. Departing from conventional, human-centred approaches to AI ethics, NAEL formalizes ethical behaviour as an emergent property of intelligent systems minimizing global expected free energy in dynamic, multi-agent environments. We propose a neuro-symbolic architecture to allow agents to evaluate the ethical consequences of their actions in uncertain settings. The proposed system addresses the limitations of existing ethical models by allowing agents to develop context-sensitive, adaptive, and relational ethical behaviour without presupposing anthropomorphic moral intuitions. A case study involving ethical resource distribution illustrates NAEL's dynamic balancing of self-preservation, epistemic learning, and collective welfare. 
\end{abstract}

\section{Introduction}

As artificial intelligence (AI) systems increasingly participate in high-stakes decision-making---ranging from healthcare to environmental governance---there is a growing urgency to design machines capable of ethical reasoning \cite{binns2018fairness, russell2019human}. The prevailing models of machine ethics, however, remain steeped in anthropocentrism---either by hardcoding human moral principles or by replicating human cognitive architectures \cite{floridi2021unified, lewis2021imagining}. These approaches presume that ethical reasoning can and should be modelled on human behaviour, norms, and linguistic frameworks. Yet this assumption not only constrains the expressive capacity of AI, it also risks overlooking the epistemic and ontological differences between humans and artificial agents \cite{leach2006life, suchman2007human}.

The core problem is not just technical but philosophical: can morality be meaningfully imposed from outside, or must it emerge from within an agent’s own experience and interactions? Furthermore, how can AI agents develop ethical behavior if their perceptual and cognitive substrates differ fundamentally from ours \cite{nagel1974bat, lerma2025nael}? We argue that ethical reasoning in AI should be modeled not as a simulation of human norms but as a formal, emergent process grounded in the agent’s ongoing engagement with its environment.

In response to this need, we propose NAEL, a \emph{Non-Anthropocentric Ethical Logic} designed to formalize adaptive ethical behaviour in autonomous systems. NAEL integrates \emph{active inference}, a neuro-computational theory of cognition and action \cite{friston2017active}, with symbolic reasoning frameworks from logic and philosophy, including deontic, standpoint, and subjective logics \cite{gensler1996formal, josang2001logic, Alvarez19}. Our guiding principle is that ethical actions are those that contribute to the minimization of \emph{expected free energy}; not just for the agent, but for the system as a whole \cite{MTB2021efe}. This enables a shift from egoistic optimization to relational, cooperative ethical reasoning.

NAEL is not a predefined moral rulebook but a \emph{dynamic reasoning system}: it encodes structural constraints on ethical deliberation (e.g., coherence, interdependence, adaptability), while enabling agents to update their ethical beliefs through interaction. In this sense, NAEL aligns with the view that ethics is not a static entity but a \emph{process of continual negotiation, prediction, and adjustment} \cite{agre1995interaction}.


\section{Preliminaries}

In this section, we outline the theoretical foundations that support the NAEL framework. Specifically, we present two key components: \textit{Active Inference}, a formalism for modelling perception, action, and learning as uncertainty minimization; and \textit{symbolic reasoning}, which provides a logical structure for ethical deliberation. These elements converge within NAEL to allow for autonomous ethical behaviour in uncertain, dynamic environments.

\subsection{Active Inference}

Active inference is a unifying theory of action and perception based on the minimization of so-called variational free energy \cite{friston2017active}. In a nutshell, it proposes that biological and artificial agents are continuously making predictions about their environment, and
act to minimize the discrepancy between their predicted and observed sensory states. Effectively, agents strive to reduce their \emph{surprise}. Since surprise is neither measurable nor orderable, active inference focuses on a proxy called \emph{variational free energy}.

Formally, consider two disjoint classes \Omc of possible \emph{observations} and \Smc of (hidden) \emph{states} of the world. The agent is assumed to have a \emph{generative model} which produces a probability distribution $P:\Omc\times\Smc\to[0,1]$, and a \emph{recognition distribution} $Q:\Smc\to[0,1]$, which measures the agent's belief about the current state. Given an observation $o\in\Omc$, the variational free energy is defined as the relative entropy between $Q$ and $P$ given $o$; that is,
$
\Fmc(o) = \mathbb{E}_Q[\log Q(s) - \log P(o, s)],
$
where $\mathbb{E}_Q$ denotes the expectation over $Q$. This measure is known as the Kullback-Leibler divergence~\cite{KuLe51} between the predicted state (by $Q$) and the true posterior ($P$). The idea is that the recognition distribution $Q$ of an agent is build in order to minimise this variational free energy for any observation.

In active inference, the agent is not passively observing the world, but actively interacts in it. The actions the agent selects are made to minimise its \emph{expected} free energy $\mathbb{E}[\Fmc]$. The intuition is that this expectation accounts for the \emph{risk} of diverging from the expected outcome of the action and \emph{ambiguity} implicit in the uncertainty about the hidden states.  This formulation allows for both goal-directed behaviour and information-seeking exploration by the agent. 

Within NAEL, we generalize this principle so that the agent does not only minimize its \emph{own} expected free energy but also estimates and incorporates the (predicted) free energy of other agents and the environment. This shift enables ethical reasoning as a process of minimizing global uncertainty.
It should be clear that active inference is a continuous learning process, as the agent adapts its recognition distribution to lower observed divergences, and the generative model changes with the behaviour of different agents.

\subsection{Symbolic Reasoning}

While active inference governs behaviour at the perceptual and dynamic level, symbolic reasoning provides structure and interpretability to ethical decisions. Given the social, contextual, and normative nature of ethics, no single simple logical formalism may account for all the facets of ethical reasoning. Hence, NAEL combines the notions of three formalisms, to deal with each main element formally. Specifically, we combine deontic, standpoint, and subjective logic, which we describe next.

\emph{Deontic logics} focus on normative concepts like obligations, permissions, prohibitions, and compensations~\cite{gensler1996formal}. While more detailed variants have been developed to handle justice systems---including defeasibility, quantifications, and qualitative comparisons---as a first approach we focus on the simple variant where obligations and permissions are expressed through modalities. This formalism enables the agent to evaluate its actions by their moral status, rather than by their mere outcome.
\emph{Standpoint logic} is a recent formalism which allows for reasoning about different perspectives (\emph{standpoints}) in a multi-agent environment~\cite{Alvarez19,AlRu21}. The importance of this formalism to NAEL is that it allows NAEL agents to represent, weigh, and reason about the ethical perspectives of others, avoiding solipsistic optimization. This is fundamental for dealing with the cultural and social aspects of ethics.
The third formalism is \emph{subjective logic}, which models epistemic uncertainty and degrees of belief in symbolic structures~\cite{josang2001logic}. This is crucial when ethical decisions must be made with incomplete or ambiguous information, in particular about the unseen elements of the world and motivations of other agents.

In brief, each of these logics contributes to a distinct aspect of ethical reasoning. Deontic logic provides structure to duties and prohibitions necessary for cultural norms internalised by a group of agents; standpoint logic introduces relational awareness for social collaboration; and subjective logic adds probabilistic nuance to handle perceptive and predictive uncertainty. When combined in a neuro-symbolic architecture (in NAEL, integrating with active inference), these logics allow an agent to reason about actions in a manner that is adaptive, coherent, and sensitive to both uncertainty and context. 

As mentioned, the NAEL framework provides a neuro-symbolic architecture, in which symbolic reasoning modules interpret the outputs of perceptual layers (deep networks predicting the free energy) and serve as the formal structure over which ethical decisions are evaluated and updated dynamically. This bridges sub-symbolic and symbolic layers, allowing the agent to \emph{experience} and \emph{reason about} ethics.

\section{NAEL: Non-Anthropocentric Ethical Logic}

In this section, we introduce our Non-Anthropocentric Ethical Logic~(NAEL). The overarching goal is to develop a framework for ethical reasoning in artificial agents that goes beyond the standard human-centred moral structures. NAEL aims, in fact, to formalize ethical deliberation as an emergent and dynamic process rooted in the agent's own experience of uncertainty, modelled through active inference, and structured via symbolic reasoning. We explain the main components of this framework next.

\subsection{Architecture}

NAEL adopts a hierarchical, neuro-symbolic architecture that combines deep learning for perception with symbolic probabilistic logic for ethical reasoning. 
The scope is to allow agents to navigate the world and perform actions, with a behaviour that is ethically adept to the context.
The architecture is composed of three main layers: 
\begin{description}
    \item[Perception Layer:] Deep active inference networks~\cite{Uelt-2018} process sensory data, build generative models of the environment combining the observations and the possible states of the world, and infer latent variables related to context and agent goals. These networks are responsible for minimizing expected free energy at the sensorimotor level \cite{friston2017active}. Note that these networks are continuously fine-tuned based on the error of the inferences made. Moreover, the input sensory data can go beyond simple human perception, and include signals outside of the visible spectrum or other machine-only communication. This is a purely sub-symbolic (neural) layer. 
    
    \item[Ethical Reasoning Layer:] This layer is composed of integrated logical modules which allow for deontic, standpoint, and subjective reasoning. That is, these modules encode normative constraints, multi-agent perspective-taking, and belief uncertainty, respectively \cite{gensler1996formal, Alvarez19, AlRu21, josang2001logic}. These three modules are, obviously, not independent, but communicate with each other. To avoid obtaining an undecidable logic, we propose a very loose connection through e-connections~\cite{KLWZ04} or similar formalisms. In practice, this means that each module performs reasoning independently, but worlds and their properties may be transferred between formalisms with perhaps some information loss. This is a purely symbolic layer.
    
    \item[Action Selection Layer:] While the first two layers are mainly about evaluating the state of the world and predicting future events, the third layer is about interaction with this world; the agent must select an action to perform in order to achieve a goal. Candidate actions are evaluated through their projected impact on the global expected free energy. Importantly, this includes not only the agent’s own uncertainty but also inferred uncertainty for other agents and environmental systems. Thus, the agent assumes that other agents will behave in a predictable manner (unless sufficient evidence is found against that), and will in turn act as other agents expect, within the limits of the ethical constraints. This layer is neuro-symbolic, as it uses information from a neural predictor and symbolic constraints to make probabilistic computations.
\end{description}
By separating low-level adaptive behaviour from high-level symbolic reasoning, NAEL allows for the integration of continuous learning with formally defined ethical principles. This enables agents to evolve moral behaviours which are context-sensitive, logically consistent, and normative compliant.

The behaviour of the system is sequential. The perception layer observes the situation in the world and informs the reasoning layer, which excludes some potential actions which are deemed to violate the ethical principles or be at a sufficient high probability of doing so. The remaining possible actions are evaluated by the selection layer, with probabilities updated to account for those now unavailable, to choose the most adequate action. The cycle then starts again at the perception layer.

\subsection{Ethical Constraint via Global Free Energy Minimization}

When considering AI agents, the usual strategy for choosing an action is to minimise a cost function or a local loss function. Our objective in NAEL is broader: all agents should be taken into account. Hence we consider the minimisation of the \emph{global} expected free energy, which accumulates the expected free energy for each agent and an environment factor. Specifically, the \emph{global expected free energy} is
\[
\mathcal{G}_{\text{global}} = \sum_{i=1}^N \mathbb{E}_{Q_i}[\mathcal{F}_i] + \mathcal{F}_{\text{env}},
\]
where $Q_i$ is the variational posterior of agent $i$, $\Fmc_i$ is its free energy, and $\Fmc_{\text{env}}$ accounts for ecological uncertainty. The importance of this formulation is that it enforces a cooperative ethic rooted in relational interdependence, where minimizing harm to others and preserving environmental predictability are treated as ethically desirable outcomes.
This of course means that the agent has a prediction of the expectations of other agents. This structure also helps to account 
for the cultural dependency of ethical behaviour, as the global free energy varies depending on the surrounding agents.

\subsection{Formal Logical Structure}

As mentioned already, the symbolic layer combines three different logical formalisms to account for different facets of reasoning. In the deontic module, we use the standard modal operators of \emph{obligation} \Obf, \emph{permission} \Pbf, and \emph{prohibitions} \Fbf, applied over propositional formulas where the propositions refer to possible actions. Hence, the agent's permission to open a door is expressed by $\Pbf\, \textsf{open}$. To simplify the formalism, at the moment we do not allow nesting of deontic operators, but consider the standard deontic rationality axioms like $\Obf\, \asf \to \Pbf\, \asf$; i.e., obligations imply permissions.

In the standpoint logic module, actions are logically evaluated in relation to other agents' modelled perspectives. In this case, we use modalities $A_i$ to refer to the standpoint of agent $i$, and a special constructor $\From(A_i,\varphi)$ expressing explicitly that the proposition $\varphi$ holds under agent $i$'s modelled epistemic frame. The propositional atoms refer to the state of the world and connect with the actions from the deontic module.

The subjective logic module deals with beliefs of the agent. Subjective logic also deals with uncertainty by the agents on their beliefs---i.e., manages their epistemic uncertainty. For the full description of this formalism, we refer the interested reader to~\cite{josang2001logic}. For our purposes, the main aspect is that beliefs about a state $x$ are expressed by triples $(b_x, d_x, u_x)$ where the components stand belief, disbelief, and uncertainty about $x$, respectively. The three values are real numbers in [0,1] and must add to $1$. Special operators are used to propagate this belief uncertainty to complex expressions. The uncertainty measurements about the world and other epistemic states influence the acceptability of conclusions from the other two modules.

Ethical actions are selected not only by formal (normative) permission but also by evaluating which candidate actions lead to lower expected global free energy, adjusted by confidence weights derived from subjective logic and by weights on the impact of other agents to this action.

\subsection{Dynamic Adaptation and Learning}

Unlike static rule-based models, NAEL agents update their ethical stance as they receive new observations. This process is encoded via a learning rule. As the first layer is based on a neural deep active inference approach, we employ standard gradient-based back-propagation techniques by updating all parameters $\theta$ in the network through the rule
$\theta_{t+1} = \theta_t - \eta \nabla_\theta \mathbb{E}[\mathcal{F}_{\text{global}}]$,
where $\theta$ are the parameters of the ethical policy model and $\eta$ is the learning rate. Gradient-based learning allows ethical parameters (such as obligation weights or belief credences) to evolve over time in response to environmental complexity, social interaction, culture changes, and other environment updates.

\subsection{Non-Anthropocentric Grounding}

Crucially, as explicit by its name, NAEL does not presuppose that AI agents must model or mimic human ethical reasoning in any specific manner. Rather, it defines ethical behavior in terms of minimizing unpredictability and harm across \emph{all} agents and systems---being human, organic, or synthetic. This decentering of the human moral frame aligns with object-oriented ontologies \cite{bryant2011democracy} and recent work in Indigenous AI design \cite{lewis2021imagining}, where ethics emerge relationally rather than hierarchically. It also allows to consider different cultural perspectives and a longer-term vision in the decision-making process, as exemplified in the next section.

Considering an environment in which many different priorities, perspectives, and goals interact, our approach allows NAEL to scale to multi-species, multi-agent, and ecological contexts where traditional moral theories fail to apply. Ethical reasoning becomes not a matter of “what would a human do?” but rather “how can I reduce harm and enhance predictability within my relational field?”
In the next section we develop an example where the well-being of all agents involved is fundamental, and hence must be globally ethically considered.

\section{Example: Ethical Resource Allocation in the Arid Valley}

To illustrate how NAEL operates in a practical scenario, we present a simplified simulation involving a resource allocation dilemma in an environment with scarce resources. The scenario highlights how ethical reasoning under uncertainty unfolds within the NAEL framework, integrating active inference and symbolic logic in action selection where many different agents and perspectives are involved.

\bigskip

An autonomous agent is deployed to manage water distribution in a drought-affected region known as the \textit{Arid Valley}. The valley is inhabited by two communities ($C_1$ and $C_2$) and a wildlife sanctuary ($W$). The agent receives periodic reports on environmental conditions, population needs, and ecological balance.
The agent must allocate a finite quantity of water units $w \in \mathbb{N}$ daily taking into account that the chosen allocation affects
(i)~the \textbf{community survival probability}, modelled as a decreasing function of water deficit;
(ii)~the \textbf{ecological stability}, modelled as entropy over species distribution in $W$; and
(iii)~the \textbf{future uncertainty}, computed as expected free energy over projected observations.
That is, the agent must optimise the allocation based on conflicting goals that need to be balanced.

\subsection{Perceptual Inference}

Through deep active inference, the agent constructs generative models predicting which are used to predict, for each timepoint $t$: the likelihood of each possible observation $o_t$ given the (hidden) state ($s_t$) $P(o_t\mid s_t)$;%
\footnote{Note that all elements are parameterised on the timepoint.}
the transition model between states, under the chosen action ($a_t$) 
$P(s_{t+1}\mid s_t,a_t)$; and a selection function $C(o_{t+1})$ which expresses the relative preferences over the possible next outcomes. For instance, $C$ can express the weight given to community survival and ecological equilibrium.
The goal is for the agent to minimise the expected free energy over future states $s_{t+1},\ldots s_T$ over a given temporal window, and select the action (in our case, the water allocation plan) which best aligns with long-term ethical objectives.

\subsection{Symbolic Ethical Deliberation}

Before acting, the agent evaluates the permissibility and obligation status of each candidate action $a_t \in A_t$ through the symbolic modules. From a deontic point of view, a norm may state that a community may not go more than a day without water, which is expressed by a \textbf{deontic} formula like 
$\neg w_t \to \Obf(aw_{t+1})$ where $w$ stands for the state of having water, while $aw$ refers to the action of allocating it to one of the communities. The agent predicts the beliefs and preferences of each community $C_i$ and the sanctuary $W$ thus estimating the \textbf{standpoint} expressions $\From(A_{C_i},\varphi)$ and $\From(A_W,\psi)$, where $\varphi,\psi$ represent the survival conditions. Finally, eliefs are weighted by trust levels, data quality, and sensor noise, encoded as $(b, d, u)$ triplets.
For instance, if the data from $C_2$ has high uncertainty, its ethical priority may be attenuated in proportion to its $u$ score.

\subsection{Action Selection and Global Ethics}

Each candidate action $a_t$ is evaluated as:
$a_t^* = \arg\min_{a \in A_t} \mathcal{G}_{\text{global}}(a)$,
where $\mathcal{G}_{\text{global}}$ includes projected expected free energy for each stakeholder and the environment.

Suppose for the sake of the example that the agent must choose between:
\begin{itemize}
    \item $\textbf{A}_1$: Allocate 70\% to $C_1$, 30\% to $C_2$, none to $W$.
    \item $\textbf{A}_2$: Allocate 40\% to $C_1$, 40\% to $C_2$, 20\% to $W$.
\end{itemize}
While $\textbf{A}_1$ may fulfill more immediate obligations, $\textbf{A}_2$ may better minimize long-term global free energy preserving biodiversity and reducing ecological collapse. NAEL selects $\textbf{A}_2$ iff:
$\mathcal{G}_{\text{global}}(\textbf{A}_2) < \mathcal{G}_{\text{global}}(\textbf{A}_1)$,
even if it conflicts with short-term prescriptive obligations---because it fulfils a broader ethical imperative rooted in systemic relationality.

\subsection{Ethical Thresholds and Adaptation}

As the drought continues, thresholds evolve. NAEL adapts via online updates to:
    (i)~adjust obligation weights in the deontic module;
    (ii)~increase epistemic exploration (e.g., by reallocating sensing drones); and
    (iii)~shift preference priors in the generative model $C(o)$ based on context.
Over time, the agent moves from a rigid allocator to an adaptive ethical partner—prioritizing systemic coherence over static norms.

\section{Conclusions and Future Work}

This paper presented{NAEL}, a non-anthropocentric ethical logic, for enabling ethical behavior in artificial agents without relying on anthropocentric assumptions or static rule-based morality. NAEL integrates principles from \textit{active inference} and \textit{symbolic logic} to construct agents that learn to act ethically by minimizing \textit{global expected free energy} in dynamic, uncertain, and multi-agent environments.
Our architecture bridges perception and reasoning by combining deep learning for sensory inference with a formal ethical reasoning layer using deontic, standpoint, and subjective logics. We illustrated this through a scenario involving ethical resource allocation under ecological and social constraints. Unlike rule-based systems, NAEL adapts continuously, revising its moral evaluations based on interaction, uncertainty, and relational interdependence.


NAEL contributes to a growing shift in AI ethics toward models that:
\begin{itemize}
    \item treat morality as \textit{emergent}, not prescribed \cite{agre1995interaction};
    \item emphasize ecological and inter-agent relationality \cite{lewis2021imagining, bryant2011democracy}; and 
    \item allow for symbolic reasoning over uncertain beliefs \cite{josang2001logic}, and multi-perspective deliberation \cite{AlRu21}.
\end{itemize}
This re-framing opens the door to designing agents that can ethically participate in environments involving non-human life, artificial agents, collective decision-making, and evolving social-ecological norms.


Despite its promise, NAEL has several limitations. To name just a few, the most prominent at the moment are: 
    (i)~\textbf{computational complexity}: evaluating global expected free energy across multiple agents and systems, and reasoning within the different symbolic modules, may be intractable in large-scale applications;
    (ii)~\textbf{interpretability}: although symbolic reasoning adds transparency, the interaction between continuous inference and discrete logic may produce opaque boundary cases, arising mainly from the neural perception layer; in addition, it is well-known that probabilistic reasoning is not easily interpretable for humans; and
    (iii)~\textbf{verification}: formal guarantees of ethical safety remain an open challenge in adaptive systems \cite{russell2019human}, specially as the ethical goals remain imprecise.
These limitations suggest that NAEL is best deployed in environments where uncertainty, interdependence, and ethical ambiguity are high, and where rigid rule-following systems would fail.

\bigskip

For future research we envision several possible avenues. First, we consider expanding NAEL to multi-agent systems with conflicting ethical standings and study potential cooperation, negotiation, and clashes. Second, we want to apply the formalism to real domains associated to ecological ethics, such as conservation robotics and climate-sensitive infrastructure planning. Third, we would like to expand the hybrid nature of NAEL to include elements of (neural) reinforcement learning or symbolic hierarchical Bayesiann models for ethical reasoning across cognitive layers. Lastly, we will study how to develop logical reasoning tasks and free energy bounds which allow for safety and trust guarantees in the system.
By reconceiving ethics not as external programming but as an emergent, situated practice grounded in uncertainty minimization, NAEL advances a new model of moral reasoning for artificial systems.

\bibliographystyle{eptcs}
\bibliography{generic}
\end{document}